%% file: main.tex
\documentclass[10pt,twocolumn,letterpaper]{article}

\usepackage[pagenumbers]{cvpr} 

\input{preamble}

\definecolor{cvprblue}{rgb}{0.21,0.49,0.74}
\usepackage[pagebackref,breaklinks,colorlinks,allcolors=cvprblue]{hyperref}

\def\paperID{25} 
\def\confName{MMFM-BIOMED @ CVPR 2026}
\def\confYear{2026}

\title{
Does Language Shift Break Medical Vision-Language Models? Indonesian Radiology Visual Question Answering Case Study
}

\author{
Pieter Christy Yan Yudhistira\textsuperscript{1} \hspace{2em}
Dzaki Rafif Malik\textsuperscript{2} \hspace{2em}
Novanto Yudistira\textsuperscript{3}\\
Intelligent System Laboratory, Faculty of Computer Science\\
Brawijaya University, Malang, Indonesia\\
{\tt\small
\textsuperscript{1}pityudhistira28@student.ub.ac.id,
\textsuperscript{2}malikdzaki16@gmail.com,
\textsuperscript{3}yudistira@ub.ac.id
}
}

\begin{document}
\maketitle
\input{sec/0_abstract}    
\input{sec/1_intro}

\input{sec/2_methods}
\input{sec/3_experiments}
\input{sec/4_conclusion}
{
    \small
    \bibliographystyle{ieeenat_fullname}
    \bibliography{main}
}


\end{document}

%% file: preamble.tex
\usepackage{graphicx}
\usepackage{float}
\usepackage{dblfloatfix}

%% file: sec/0_abstract.tex
\begin{abstract}


Medical Vision–Language Models (VLMs) are typically evaluated on English radiology visual question answering benchmarks, leaving their robustness under non-English clinical language largely unexplored. We introduce \textbf{IndoRad-VQA}, an Indonesian adaptation of VQA-RAD, to assess whether medical VLMs retain radiology reasoning ability when questions are asked in Bahasa Indonesia. Radiology question–answer pairs are translated into Indonesian with self-evaluation–based quality control to preserve clinical meaning, terminology consistency, and answer equivalence. We evaluate general-purpose, Southeast Asian multilingual, and medical-specific VLMs under English and Indonesian prompting settings. Beyond accuracy, we also quantify the language robustness gap between English and Indonesian inputs. We also conduct an error analysis to identify failure modes of question-answering, such as yes/no flips, laterality errors, and output-language mismatches. Our findings show that strong performance on English medical VQA benchmarks does not necessarily translate to robust behavior in Indonesian clinical contexts. We observe a performance gap of 8–25\% between the English and Indonesian settings, depending on the evaluation metric. This highlights the need for more inclusive multilingual evaluation of medical multimodal foundation models. The dataset are available at \href{https://huggingface.co/datasets/Lab-IS/IndoRad-VQA}{this https URL}.
\end{abstract}

%% file: sec/1_intro.tex
\section{Introduction}
\label{sec:intro}
Radiology visual question answering (VQA) has emerged as a key evaluation of medical capabilities for Vision Language Models (VLMs)~\cite{lau2018vqa_rad} \cite{liu2021slake} \cite{butsanets2025radimagenet_vqa}. Benchmarks such as VQA-RAD~\cite{lau2018vqa_rad} and SLAKE~\cite{liu2021slake} measure the model's ability to answer a clinically grounded question about a radiological scan or internal human anatomy. However, most established benchmarks are constructed exclusively in English, with the non-English benchmarks often have fewer question-answer pairs compared to English counterpart, with the exception of the newer VQA benchmark~\cite{milut_japanese_medical_vqa_12m}~\cite{Ding2025MedBenchVA}. This creates a significant evaluation gap for the majority of the world's population who access medical care in other languages.

Bahasa Indonesia is spoken by over 270 million people and is the primary language of medical practice across Indonesian hospitals. Yet there is currently no dedicated Indonesian-language benchmark for evaluating radiology VQA. This absence means that clinical deployment and evaluation for VLMs are made without evidence of robustness in the target operational language in Indonesia. 

We address this gap by introducing \textbf{IndoRad-VQA}, an Indonesian language adaptation of established English radiology VQA. Our key insight is that translating the \emph{question} while keeping the \emph{image} fixed provides a controlled testbed for isolating language-induced
failures from visual reasoning failures. A model that correctly answers an English question but fails on the semantically identical Indonesian question exposes a language robustness deficit, indicating that language shift alone can impact the model's clinical reasoning.

\noindent \textbf{Research question.} \textit{Do medical VLMs that perform well on English radiology VQA preserve their visual reasoning capability when clinical questions are posed in Bahasa Indonesia?}

\noindent \textbf{Contributions.} We make four main contributions:
\begin{enumerate}
    \item We introduce \textbf{IndoRad-VQA}, an Indonesian evaluation set derived from an established English VQA benchmark \textbf{with self-evaluation quality control for terminology and answer equivalence.}
    \item We propose a \textbf{bilingual evaluation protocol} measuring strict accuracy, normalized accuracy, F1 tokenized, and BERT score, and a Language Robustness Gap (LRG) metric.
    \item We \textbf{benchmark} seven \textbf{open-source VLMs} spanning general-purpose, Southeast Asian multilingual, and medical-specific models across English and Indonesian settings.
    \item We provide a \textbf{failure-mode taxonomy} for language-induced errors with qualitative examples and an error distribution analysis.
\end{enumerate}

\noindent \textbf{Scope.} This work is a benchmark and evaluation study. The results should not be interpreted as evidence of clinical deployment readiness.

%% file: sec/2_methods.tex
\section{IndoRad-VQA: Dataset and Protocol}
\label{sec:method}

\subsection{Source Dataset}
We construct our benchmark based on VQA-RAD~\cite{lau2018vqa_rad}, a radiology visual question answering dataset that includes 2,248 question–answer pairs over 315 medical images. The dataset covers three primary imaging modalities: 104 head axial single-slice CT/MRI scans, 107 chest X-rays, and 104 abdominal axial CT scans.

\subsection{Translation Pipeline}
The translation pipeline is inspired by the proposed pipeline for Anak Baik~\cite{hakim-etal-2025-anak}, a curated set of ethical and unethical instructions derived from an established English benchmark.

\begin{figure}[ht]
    \centering
    \includegraphics[width=0.9\linewidth]{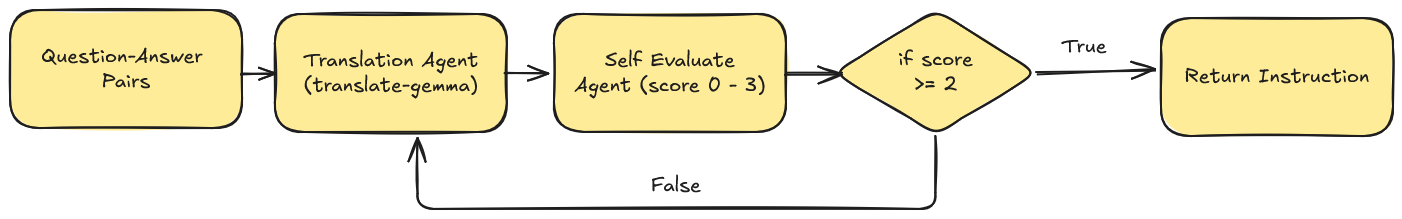}
    \caption{Self-evaluated translation pipeline for VQA-RAD English–Indonesian conversion.}
    \label{fig:translation-pipeline}
\end{figure}

\noindent \textbf{Step 1 — Machine translation.} We translate all English question and
answer strings to Indonesian using a translategemma-4b-it~\cite{Finkelstein2026TranslateGemmaTR}, an open-source model on Hugging Face, prompting it to preserve
medical terminology where no standard Indonesian equivalent exists.

\noindent \textbf{Step 2 — Automated cleaning.} We apply lower-casing, whitespace
normalization, and explicit mapping of binary pairs: \textit{yes}/\textit{ya},
\textit{no}/\textit{tidak}, \textit{right}/\textit{kanan}, \textit{left}/\textit{kiri}.

\noindent \textbf{Dataset schema.} The final dataset includes columns:
\texttt{image\_id}, \texttt{question\_en}, \texttt{answer\_en}, \texttt{question\_id},
\texttt{answer\_id}, \texttt{answer\_type}, \texttt{question\_type}, \texttt{split}.

\subsection{Answer Normalization}
Multilingual evaluation is prone to false penalization~\cite{wieting-etal-2019-beyond}\cite{Licht2022ConsistentHE}. This occurs where a model outputting a synonymous or semantically correct output, but fails under exact match evaluation. We construct a bilingual (Indonesia-English) normalization dictionary
(Table~\ref{tab:norm}) that applied before all accuracy evaluation. 

\begin{table}[ht]
\small
\centering
\caption{Bilingual answer normalization groups (extended with anatomy/radiology terms).}
\label{tab:norm}
\begin{tabular}{p{2.2cm}p{5.3cm}}
\hline
\textbf{Equivalence Group} & \textbf{Accepted Variants} \\
\hline
Yes & yes, ya, iya, iyaa, benar, betul, ada, positive, positif, yep, yeah \\
No & no, tidak, tdk, bukan, tidak ada, nope, negative, negatif \\
Right & right, kanan, sisi kanan, right side \\
Left & left, kiri, sisi kiri, left side \\
Bilateral & bilateral, kedua sisi, kiri dan kanan, kanan dan kiri, both sides, dua sisi \\
Frontal Lobe & frontal lobe, lobus frontal \\
Heart Border & heart border, batas jantung \\
Corpus Callosum & corpus callosum, korpus kalosum \\
Consolidation & consolidation, konsolidasi, konsolidasi ruang udara, airspace consolidation \\
Infarct & infarct, infark \\
Lung & lung, paru, paru-paru \\
Liver & liver, hati, hepar \\
Kidney & kidney, ginjal, ren \\
Brain & brain, otak \\
\hline
\end{tabular}
\end{table}

\subsection{Evaluation Settings}
For the evaluation, we define two controlled settings:
\begin{itemize}
    \item \textbf{EN-original}: original VQA-RAD English questions as baseline.
    \item \textbf{ID-translated}: Indonesian questions, with instruction in Bahasa Indonesia. 
\end{itemize}

\subsection{Metrics}
We use five complementary metrics for model's evaluation:
\begin{itemize}
    \item \textbf{Strict Accuracy}: Exact match after lowercase and whitespace trimming.
    \item \textbf{Normalized Accuracy}: Exact match after bilingual normalization, lowercase and whitespace trimming.
    \item \textbf{Tokenized F1}: Computes the mean of precision and recall based on the overlap between predicted tokens and ground-truth (human-annotated) tokens.
    \item \textbf{BERT-Score}~\cite{Zhang2019BERTScoreET}: Use pre-trained contextual embeddings from google-bert/bert-base-multilingual-cased~\cite{DBLP:journals/corr/abs-1810-04805} and matches output sentences by cosine similarity. 
    \item \textbf{Language Robustness Gap (LRG)}: $\text{Acc(EN)} - \text{Acc(ID)}$;
          positive values indicate language-induced performance degradation.
\end{itemize}
Results are further broken down by question type of closed (yes/no) or open questions. 

\subsection{Models}
We evaluate seven models spanning three categories:
\begin{itemize}
    \item \textbf{General-purpose VLMs}: Qwen3-VL-8B-Instruct~\cite{qwen3technicalreport}, InternVL3-2B~\cite{Zhu2025InternVL3EA}.
    \item \textbf{SEA-multilingual VLMs}: Gemma-SEA-LION-v4-8B-VL~\cite{Ng2025SEALIONSA}, Qwen-SEA-LION-v4-4B-VL~\cite{Ng2025SEALIONSA}, Qwen-SEA-LION-v4-8B-VL~\cite{Ng2025SEALIONSA}. 
    \item \textbf{Medical-specific VLMs}: MedVLM-R1~\cite{pan2025medvlm}, MedGemma-v1.5-4B~\cite{Sellergren2026MedGemma1T}.
\end{itemize}

All models receive the same image, the same question-answers pair and a standardized zero-shot prompt per setting. No fine-tuning is performed in this study.

%% file: sec/3_experiments.tex
\begin{table*}[!ht]
\small
\centering
\caption{Main results on the VQA-RAD all-set (2,248 QA pairs).
EN = English (strict), EN* = English (normalized),
ID = Indonesian (strict), ID* = Indonesian (normalized).
F1 = F1 Tokenized, F1* = F1 Tokenized (normalized), BERT = BERT Score using google-bert/bert-base-multilingual-cased~\cite{DBLP:journals/corr/abs-1810-04805}. BERT* = BERT Score (normalized).
GEN = General Model, SEA = SEA-Multilingual Model, MED = Medical Model. $\downarrow$ indicates model's degradations on Indonesian responses. \underline{underline} indicates highest performance across all models.
}
\label{tab:main}
\begin{tabular}{lccccccccc}
\hline
\textbf{Model} & \textbf{Type} & \textbf{EN} & \textbf{ID} & \textbf{EN*} & \textbf{ID*} & \textbf{F1 EN} & \textbf{F1 ID*} & \textbf{BERT EN} & \textbf{BERT ID*}\\
\hline
Qwen3-VL-8B-Instruct & GEN & \underline{51.02} & $\downarrow$16.00 & \underline{51.11} & $\downarrow$40.29 & \underline{56.11} & $\downarrow$44.88 & 57.40 & $\downarrow$43.85 \\

InternVL3-2B & GEN & 41.00 & $\downarrow$25.40 & 41.00 & $\downarrow$29.77 & 41.53 & $\downarrow$30.88 & 41.09 & $\downarrow$38.90 \\

Gemma-SEA-LION-v4-4B-VL & SEA & 40.20 & $\downarrow$21.57 & 40.42 & $\downarrow$36.90 & 45.99 & $\downarrow$42.07 & 63.99 & $\downarrow$\underline{54.90} \\

Qwen-SEA-LION-v4-4B-VL & SEA & 48.17 & $\downarrow$18.00 & 48.26 & $\downarrow$41.13 & 52.85 & $\downarrow$45.19 & \underline{65.20} & $\downarrow$52.23 \\

Qwen-SEA-LION-v4-8B-VL & SEA & 50.53 & $\downarrow$17.96 & 50.62 & $\downarrow$41.18 & 55.93 & $\downarrow$45.99 & 60.56 & $\downarrow$47.95 \\   

MedVLM-R1 & MED & 37.17 & $\downarrow$12.52 & 37.34 & $\downarrow$30.57 & 39.71 & $\downarrow$30.93 & 47.37 & $\downarrow$31.11 \\

MedGemma-v1.5-4B & MED & 50.62 & $\downarrow$\underline{25.45} & 50.98 & $\downarrow$\underline{44.39} & 55.93 & $\downarrow$\underline{49.20} & 52.67 & $\downarrow$45.10 \\
\hline
\end{tabular}
\end{table*}

\section{Experiments and Results}
\label{sec:experiments}

\subsection{Main Results}
Table~\ref{tab:main} reports strict accuracy, normalized accuracy, F1 tokenized, BERT score and the Language Robustness Gap (LRG) for all models across EN-original and ID-translated settings.

A substantial performance decline across all models indicates consistent degradation when radiology questions are presented in Bahasa Indonesia. This pattern is also found in medically specialized models, which despite their domain-specific training, do not exhibit robustness to language shift. These findings suggest that clinical-domain training in current open-source models is insufficient to mitigate the English-centric language bias inherent in VLMs.

\begin{table}[H]
\small
\centering
\caption{Language Robustness Evaluation and Gaps. EN = English, ID = Bahasa Indonesia. LRG = EN Acc $-$ ID Acc. $\downarrow$ model's degradations on Indonesian responses. \underline{underline} indicates highest performance across all models.
}
\label{tab:bymetric}
\begin{tabular}{lccc}
\hline
\textbf{Metric} & \textbf{EN} & \textbf{ID} & \textbf{LRG} \\
\hline
Strict         & 45.09 & $\downarrow$19.82 & 25.27 \\
Normalized    & 45.25 & $\downarrow$37.18 & 8.07 \\
F1 Tokenized    & \underline{49.20} & $\downarrow$\underline{40.66} & 8.54 \\
BERT Score    & 53.85 & $\downarrow$43.63 & 10.21 \\
\hline
\end{tabular}
\end{table}

Table~\ref{tab:bymetric} also suggests that the language shift to Bahasa Indonesia affects strict accuracy more than other evaluation metrics. The gap between overall strict accuracy and the other metrics is nearly 20\%, meaning a substantial portion of answers are marked incorrect under strict evaluation. This indicates that many model outputs may be semantically correct, but fail to exactly match the ground truth format required by the strict criterion.




\subsection{Failure-Mode Analysis}

We implement an automated error-case detection pipeline to support failure mode analysis of our results. Specifically, we analyze whether model failures are dominated by (1) incorrect responses to yes/no questions, or (2) cross-lingual inconsistency, where the model answers correctly in the original English (EN) setting but incorrectly in the translated Indonesian (ID) setting. Table~\ref{tab:errors} presents
the error taxonomy.

\begin{figure}[h]
    \centering
    \includegraphics[width=0.9\linewidth]{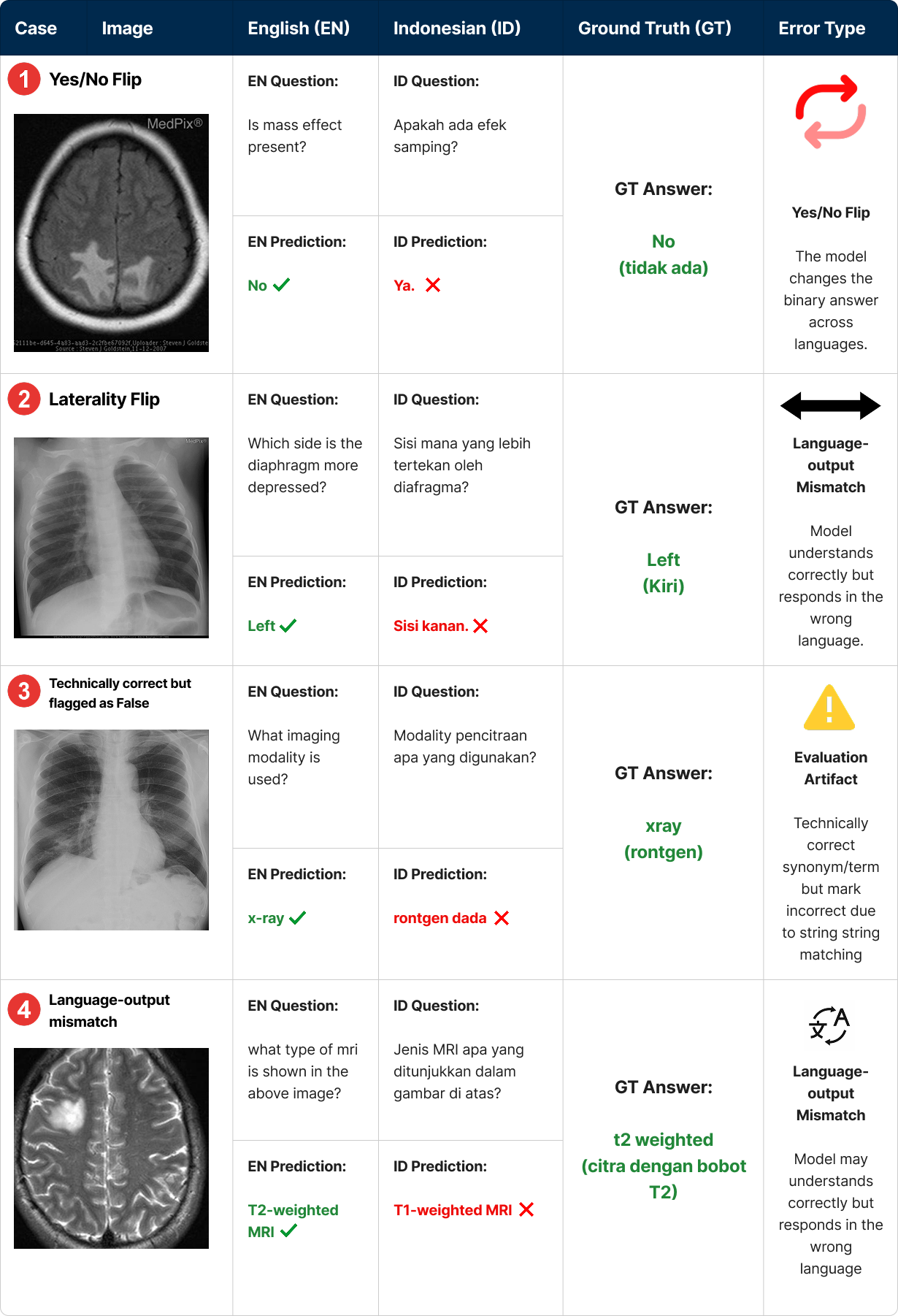}
    \caption{Qualitative failures examples of yes/no flip, language-output mismatch, laterality flip and others VLMs type of failures.}
    \label{fig:examples}
\end{figure}

Table~\ref{tab:errors} shows that other errors (terminology and visual) are the most common mistake observed in our experiments, followed by yes/no flips errors and other errors type. The errors pattern is illustrated in Figure~\ref{fig:examples}, where the model gives an incorrect yes/no, language output mismatch, and laterality flip response to a question written in Bahasa Indonesia.

\begin{table}[H]
\small
\centering
\caption{Failure-mode distribution (EN correct, ID incorrect). n=7,990 (Total number of all-set errors aggregated across all experiments).}
\label{tab:errors}
\begin{tabular}{lcc}
\hline
\textbf{Error Type} & \textbf{Count} & \textbf{\%} \\
\hline
Yes/No flip               & 1,224 & 15.3~\% \\
Laterality flip          & 18 & 0.2~\% \\
Language-output mismatch  & 89 & 1.1~\% \\
Other (terminology / visual)  & 6659 & 83.3~\% \\
\hline
\end{tabular}
\end{table}

\noindent \textbf{Qualitative examples.} Figure~\ref{fig:examples} shows
representative failure cases. For open-ended questions, we observe different types of errors, including output-language mismatches, where the model answers in English despite receiving an Indonesian prompt, as well as add-ons specific responses to questions that expect short one-word or two-word answers.



%% file: sec/4_conclusion.tex
\section{Conclusion}
\label{sec:conclusion}

We introduced \textbf{IndoRad-VQA}, an Indonesian adaptation benchmark for evaluating language robustness of medical VLMs on Indonesian radiology VQA. Our evaluation of
open-source models demonstrates a consistent language robustness gap: models that perform well on English VQA degrade substantially when the same clinical questions are posed in Bahasa Indonesia. Crucially, medical-specific pre-training does not confer language robustness. This may be a sign that the deficit is language-driven, not vision-driven. Our failure-mode analysis shows that the majority of errors fall under terminology and visual reasoning, while yes/no flips emerge as the most prominent language-induced failure category. Output-language mismatch and laterality flips occur less frequently but reveal interpretable failure patterns.

\noindent \textbf{Limitations.} Medical validation was currently performed on a single radiology VQA dataset, merging several open radiology datasets is planned for subsequent experiments. The translation was performed using a single machine translation model, TranslateGemma (4B parameters), although larger variants (12B and 27B) are also available. The 4B model was selected due to computational resource constraints. The results are limited to zero-shot evaluation. This study does not evaluate clinical decision-support readiness of VLMs. Medical validation was conducted via self-evaluation rather than radiologist review.

\noindent \textbf{Release.} We will release the IndoRad-VQA translation files,
bilingual normalization dictionary, prompt templates, and evaluation scripts to
support reproducibility of the research.